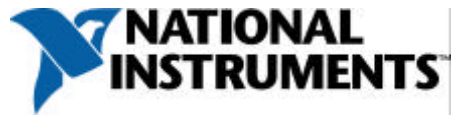

# Fingerprint based bio-starter and bio-access

**Gerardo Iovane**

**D.I.I.M.A. - University of Salerno**

**Paola Giordano**

**University of Salerno**

**Costanza Iovane and Fabrizio Rotulo**

**IXFIN S.p.A**

- Electronics

**Products used**
- LABVIEW  Prof Dev Sys 6.1
- LABVIEW  Real Time Module 6.1
- IMAQ Vision 6.01
- SQL TOOLKIT 2.0
- INTERNET TOOLKIT 5.0
- SIGNAL PROCESSING TOOLSET
- SPC Tools

**The challenge:** The development of a prototype system, based on fingerprint analysis, for people identifying and recognizing  to allow the access and to able to the starter on the car.
.

**The solution:** A real vision system is projected and implemented with innovative algorithms to execute the recognition in multidimensional spaces and transformed ones respect to the space of the native image.

## Abstract

In the paper will be presented a safety and  security system based on fingerprint technology. The results suggest a new scenario where the new cars can use a fingerprint sensor integrated in car handle to allow access and in the dashboard as starter button.

## Introduction
One of the most relevant question in the security world is how identify and recognize people not only  in some protected areas with a supervisor, but also in  public area like as in airports, banks, etc.
The automotive segment is another  example of how keys, PIN pads, and tokens have been used to provide the security mechanism for the auto. Even though many "high end" autos use a transponder key to initiate the engine ignition process, if this key is stolen or given to an unauthorized person, it will still provide access to unauthorized person, just as if he were authorized.

The most relevant parameters are the FRR (**F**alse **R**ejection **R**ate) and the FAR (**F**alse **A**cceptance **R**ate): the first one is linked with the safety while the second one with the security. Here we present a system based on the fingerprint sensor by Authentec and Labview Environment by NI, which has a scaleable accuracy respect to different sensors and so respect to the request and cost. However we deal with low cost system. In our system the FRR parameter is 1/100, while the FAR is 1/100.000 that in the present scenario is a good result and it allows us to consider a possible use for car security and safety system.

## The Fingerprint System

The hardware consists of:
- a x86 based hardware with the Microsoft Window as OS (we are waiting the release 7 of Labview to produce an embedded system based on Intel Strong Arm processor and Windows CE),
- an EntrèPad AES 3500 sensor by Authentec.

The system architecture is structured as shown in Fig.1.

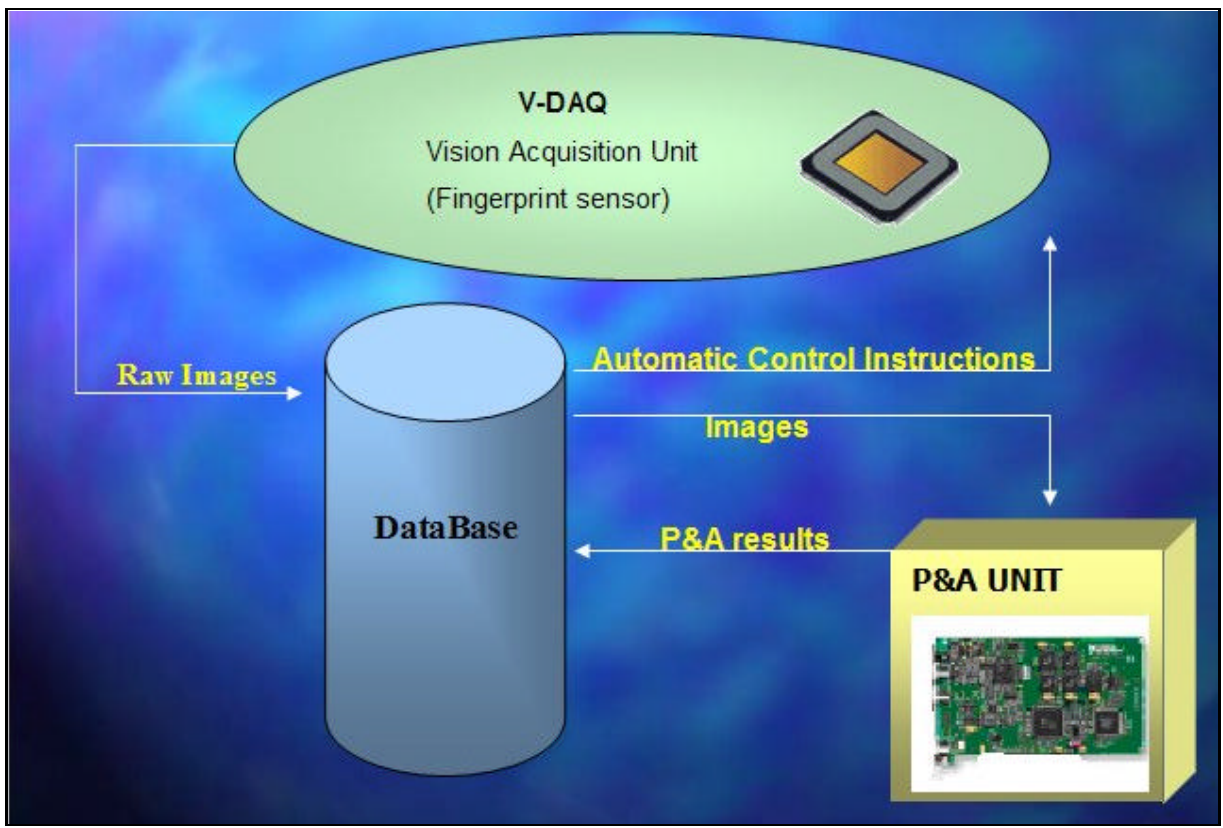

**Figure 1:** The system architecture.

The software is realized with Labview and Labview RT. The image analysis is realized with IMAQ and the data analysis use the SIGNAL PROCESSING TOOLSET by NI. A private library was developed for the recognition based on wavelet analysis. It uses SQL language, and is linked to the system by SQL TOOLKIT by NI.

Each raw image (see Fig 2) is given to the Process and Analysis Unit (P&A) for the recognition. This unit is formed by four subunits: 1) the first tests the morphology of the finger in the real space; 2) the second subunit transforms the image in a 3-d image, where the third dimension corresponds to a different weight of the finger respect to a fixed coordinate frame and respect to some characteristic parameters (see Fig.3); 3) the third subunit transforms the image in the frequencies domain where a wavelets analysis is carried out (here the best coefficients are fixed thanks to a neural network based on a multilayer perceptron); 4) the last subunit transforms the image in a multidimensional objects, where the dimensions of the space are linked with some principal parameters of the finger: curves, lines and minutiae; then the algorithm executes the last analysis to recognize a person by using a genetic pipeline.

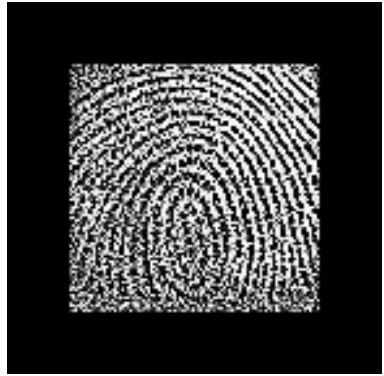

**Figure 2:** A typical fingerprint.

After the analysis the data come back the DB. In particular, statistics, plots of data and events are produced and stored by this module. In the occurrence of a special events, like as an alert, this unit can automatically reach supervisors and the police with e-mail service and SMS (Short Message System). This module is put into practice by using the INTERNET TOOLKIT and the SPC tools by NI.

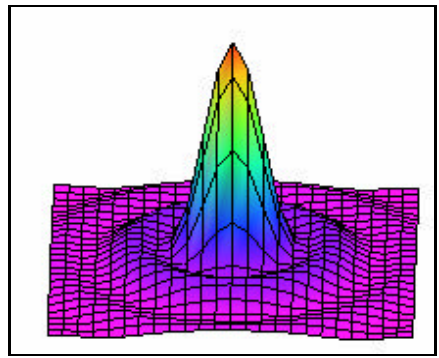

**Figure 3:** A 3-d fingerprint in a transformed space.

### Results and Benefits
By using a fingerprint authentication system, the consumer may enter and start the vehicle without the use of keys. Automotive manufacturers will also be able to provide consumers a unique combination of security, convenience and personalization. Keys, as well as insecure PIN codes or passwords, will eventually be rendered obsolete because vehicles may be activated by biometric information that cannot be lost, stolen or forgotten. Personalized levels of access may also be specified for different users of a vehicle, including:

1. Access to storage compartments

2. Access to e-mail and radio settings

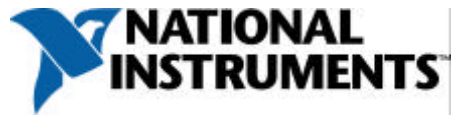

3. Seat adjustments
4. Mirror adjustments
5. Climate control settings

The Benefits of a Fingerprint Validation System are:

1. Virtually eliminates unauthorized access
2. Reduces possibility of auto theft
3. Allows personalization at the touch of a finger
4. Appeals to security-conscious consumers
5. Appeals to tech-savvy consumers

## Conclusion

An high challenge for safety and security is reach and an interesting solution is fixed. The present system get high level results in fingerprint recognition. The accuracy and the precision are appropriate for the automotive segment. Fingerprint System is scalable respect to the needed level of security and the budget. It is low cost system respect to the standard cost per year in public structure or enterprise.

## Acknowledgements

The authors wish to thank I.Piacentini and the NI imaging group in Italy for relevant software suggestions and comments.